\documentclass[10pt,conference,letterpaper]{IEEEtran}

\IEEEoverridecommandlockouts

\usepackage{cite}
\usepackage{algorithmic}
\usepackage{graphicx}
\usepackage{textcomp}
\usepackage{xcolor}
\usepackage{algorithm}
\usepackage{algorithmic}
\usepackage{graphicx}  
\usepackage[T1]{fontenc}
\usepackage{times}
\usepackage{mathptmx}   
\usepackage{helvet}
\usepackage{courier}
\IEEEoverridecommandlockouts
\IEEEpubid{\makebox[\columnwidth]{\hfill © 2025 IEEE \hfill}%
\hspace{\columnsep}\makebox[\columnwidth]{ }}
\begin{document}

\title{Federated Learning with Gramian Angular Fields for Privacy-Preserving ECG Classification on Heterogeneous IoT Devices}

\author{\IEEEauthorblockN{Youssef Elmir}
\IEEEauthorblockA{\textit{Laboratoire LITAN} \\
\textit{École supérieure en Sciences et Technologies}\\
\textit{de l’Informatique et du Numérique}\\
RN 75, Amizour 06300, Béjaia, Algérie \\
elmir@estin.dz}
\and
\IEEEauthorblockN{Yassine Himeur}
\IEEEauthorblockA{\textit{College of Engineering and IT} \\
\textit{University of Dubai}\\
Dubai, UAE \\
yhimeur@ud.ac.ae}
\and
\IEEEauthorblockN{Abbes Amira}
\IEEEauthorblockA{\textit{Department of Computer Science} \\
\textit{University of Sharjah}\\
Sharjah, UAE \\
aamira@sharjah.ac.ae}

}

\maketitle
\IEEEpubidadjcol

\begin{abstract}
This study presents a federated learning (FL) framework for privacy-preserving electrocardiogram (ECG) classification in Internet of Things (IoT) healthcare environments. By transforming 1D ECG signals into 2D Gramian Angular Field (GAF) images, the proposed approach enables efficient feature extraction through Convolutional Neural Networks (CNNs), while ensuring sensitive medical data remain local to each device. This work is among the first to experimentally validate GAF-based federated ECG classification across heterogeneous IoT devices, quantifying both performance and communication efficiency. To evaluate feasibility in realistic IoT settings, we deployed the framework across heterogeneous IoT devices including a server, a laptop, and a resource-constrained Raspberry Pi 4 reflecting edge–cloud integration in IoT ecosystems. Experimental results demonstrate that the FL-GAF model achieves a high classification accuracy of 95.18\% in a multi-client setup, significantly outperforming a single-client baseline in both accuracy and training time. Despite the added computational complexity of GAF transformations, the framework maintains efficient resource utilization and communication overhead. These findings highlight the potential of lightweight, privacy-preserving AI for IoT-based healthcare monitoring, supporting scalable and secure edge deployments in smart health systems.
\end{abstract}

\textbf{Keywords}: Federated Learning, Internet of Things, ECG Classification, Gramian Angular Field, Convolutional Neural Networks, Heterogeneous Devices, Edge Computing.

\section{Introduction}

\vspace{5pt}
\begin{center}
    \textit{© 2025 IEEE. This is the author’s version of the work accepted for publication in the IEEE Computers, Communications and IT Applications Conference (ComComAp 2025). 
    The final version will be available via IEEE Xplore.}
\end{center}
\vspace{5pt}

Federated Learning (FL) enables privacy-preserving training of machine-learning models on distributed electrocardiogram (ECG) data, allowing collaborative development without sharing raw patient records \cite{donkada2023uncovering}. Prior studies confirm that FL can produce ECG classification models for cardiovascular disease diagnosis with performance comparable to centralized approaches \cite{agrawal2024federated,jimenez2023application}. Within Internet-of-Things (IoT) healthcare systems, where distributed devices must operate securely and efficiently, FL addresses key challenges of privacy, bandwidth, and data sovereignty \cite{imteaj2021survey}.

Deploying FL on heterogeneous, resource-limited devices such as Raspberry Pi remains difficult due to limited computation, bandwidth, and energy. Nevertheless, FL reduces data-transfer needs and supports scalable cloud–edge healthcare architectures \cite{jimenez2023application}. Gao et al. \cite{gao2020end} compared FL and Split Neural Networks for IoT applications and found that FL offers superior communication efficiency and robustness to non-IID data, though SplitNN converges faster on balanced datasets. They emphasized the need for further work on energy, memory, and communication optimization to strengthen FL’s practical viability.

For ECG classification, FL enables privacy-preserving model training across diverse client environments, yet non-IID data and device heterogeneity still challenge global convergence \cite{sakib2021asynchronous,diao2020heterofl,ccelik2023comparison}. Studies on 12-lead ECG data have achieved up to 98\% accuracy in IID conditions \cite{jimenez2023application,ccelik2023comparison}, demonstrating FL’s ability to preserve privacy while maintaining diagnostic performance across devices.

Meanwhile, transforming 1D ECG signals into 2D Gramian Angular Field (GAF) images has shown improved classification accuracy in multiple works \cite{elmir2023ecg,yoon2025enhanced,yousuf2024inferior,yang2024detection,zhang2019automated}. GAF encodes temporal dynamics as spatial correlations, enabling Convolutional Neural Networks (CNNs) to extract richer features. Although not universally optimal, empirical evidence supports GAF’s effectiveness for diverse ECG analysis tasks.

However, GAF transformations increase computational and memory costs on lightweight hardware. Camara et al. \cite{camara2023ecg} and Gao et al. \cite{gao2025efficient} highlighted this trade-off and proposed optimized 1D-to-2D conversions for deployment on Raspberry Pi 4. Subsequent studies \cite{eleftheriadis2024energy,alsalemi2023lightweight} achieved acceptable latency through simplified transformations and compact model architectures, confirming that careful optimization can make GAF feasible on edge devices.

\textbf{The main contributions of this work are as follows:}
\begin{itemize}
\item We formulate a federated learning (FL) framework tailored for electrocardiogram (ECG) classification in IoT healthcare environments, integrating Gramian Angular Field (GAF) transformations with Convolutional Neural Networks (CNNs). This framework is designed to test the hypothesis that combining GAF-based spatial representations with decentralized FL training can achieve high diagnostic accuracy while preserving patient data privacy.
\item We present a reproducible experimental methodology using a heterogeneous edge–cloud configuration (server, laptop, Raspberry Pi 4). The setup quantifies the trade-offs among model accuracy, training time, communication cost, and device resource utilization, providing an evidence-based assessment of FL feasibility in constrained IoT scenarios.
\item We demonstrate that the proposed FL–GAF model achieves 95.18\% classification accuracy in a multi-client setting, significantly outperforming the single-client baseline. In addition, we analyze the effects of heterogeneous participation and non-IID data on global model convergence and discuss communication–performance trade-offs relevant to scalable, privacy-preserving healthcare deployments.
\end{itemize}

Unlike prior GAF-FL studies, this work uniquely investigates deployment feasibility on heterogeneous IoT hardware, quantifying performance–resource trade-offs and demonstrating lightweight adaptability for edge-based healthcare.

The remainder of this paper is organized as follows: Section II details the methodology, covering the FL framework, GAF transformation, and CNN architecture; Section III describes the dataset and device configurations; Section IV presents results and discussion, and Section V concludes with limitations and future research directions.

\section{Methods}
\subsection{Federated Learning Framework}
Our FL framework comprises a central server and two additional heterogeneous clients as presented in Figure \ref{fig:fw_framework}:

\begin{figure}[ht]
    \centering
    \includegraphics[width=1\linewidth]{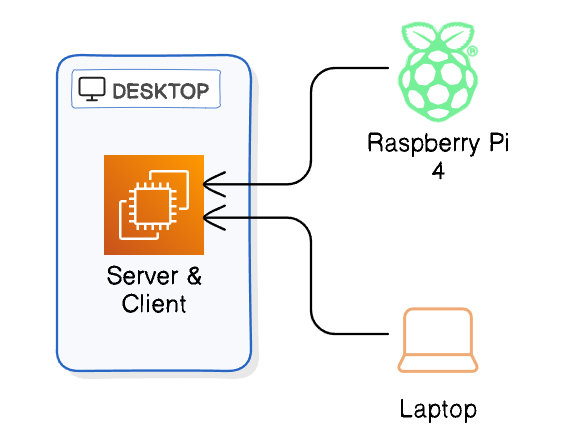}
    \caption{Proposed Federated Learning Framework}
    \label{fig:fw_framework}
\end{figure}

\begin{algorithm}
\caption{Federated Learning Process}
\begin{algorithmic}[1]
\REQUIRE Clients $C_1, C_2, \ldots, C_N$, Server $S$
\STATE Server initializes global model $\theta$
\FOR{each round $r = 1, 2, \ldots, R$}
    \STATE Server sends global model $\theta$ to each client $C_i$
    \FOR{each client $C_i$ in parallel}
        \STATE Client $C_i$ updates model locally to $\theta_i$ on its data
        \STATE Client $C_i$ sends updated model $\theta_i$ back to server
    \ENDFOR
    \STATE Server aggregates models $\theta_1, \theta_2, \ldots, \theta_N$ to update $\theta$
\ENDFOR
\RETURN Global model $\theta$
\end{algorithmic}
\label{algo:1}
\end{algorithm}

\begin{itemize}
    \item \textbf{Server}: Hosts the global model, aggregates updates from clients, and manages communication.
    \item \textbf{Clients}: In addition to the one on the server, a laptop and a Raspberry Pi 4, each with different computational resources. Each client executes local training on its subset of data and transmits model weights to the server for aggregation.
\end{itemize}

The server initiates each federated round by distributing the global model weights to the clients, as outlined in Algorithm \ref{algo:1}. Each client then performs local training on its dataset for ten epochs, updating the model weights independently. Upon completing the local training, clients send their updated weights back to the server, where the models are aggregated to form a new global model for the subsequent round.

\subsection{GAF Transformation for ECG Classification}

\begin{figure*}[ht]
\centering
\includegraphics[width=0.48\linewidth]{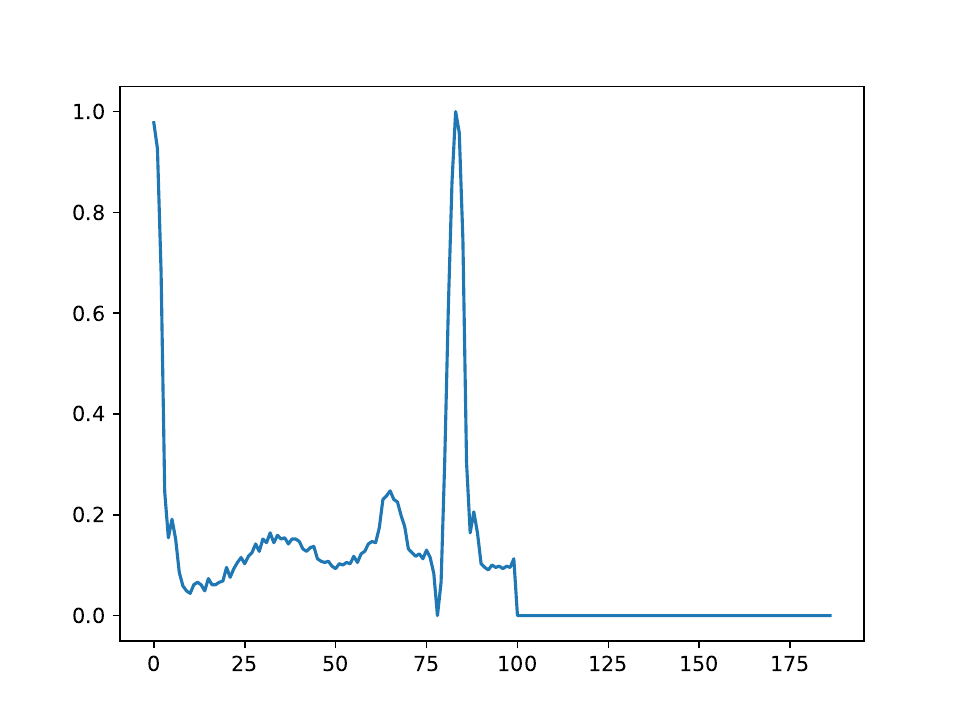}
\includegraphics[width=0.48\linewidth]{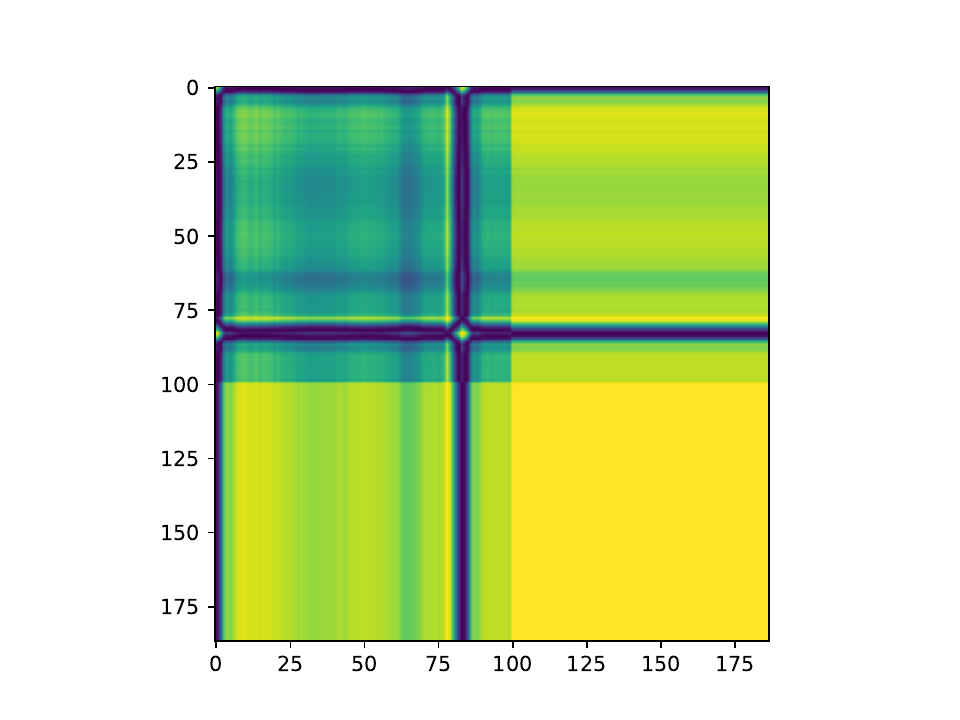}
\caption{Example of transforming a 1D ECG signal (of a sample) to a 2D GAF image: (a) 1D vector, and (b) ECG 2D GAF image\cite{elmir2023ecg}.}
\label{fig:2}
\end{figure*}

The Gramian Angular Field (GAF) transformation converts 1D time-series data, such as ECG signals, into 2D image representations by encoding each sample as an angular value in polar coordinates. This process captures temporal correlations between signal points, producing structured visual patterns that can be effectively processed by Convolutional Neural Networks (CNNs) for classification tasks \cite{elmir2023ecg}. Each element of the resulting Gramian matrix reflects the cosine of the summed angles between time-series values \cite{wang2015encoding}, allowing the model to uncover temporal dependencies and subtle variations in cardiac activity that may be less apparent in the original 1D domain.

As illustrated in Figure \ref{fig:2}, the GAF transformation converts a sample 1D ECG signal into a 2D GAF image, enabling CNNs to process ECG data in a spatial format. This facilitates improved model performance through spatial feature extraction, leveraging the 2D structure of GAF representations \cite{campanharo2011duality}.

In previous work, the GAF method demonstrated high accuracy for ECG classification, achieving up to 97.47\% accuracy and 98.65\% F1-score for anomaly detection \cite{elmir2023ecg}. These results support the viability of GAF as a feature representation approach, particularly for identifying complex ECG patterns across multiple datasets. By resizing the transformed GAF images to a uniform 32x32 size, our study ensures compatibility with the CNN model while reducing computational load, further advancing the applicability of GAF in distributed, real-time ECG analysis scenarios.

\subsection{Model Architecture}
\begin{figure*}[!htbp]
    \centering
    \includegraphics[width=0.9\linewidth, trim={200} {170} {-20} {150}, clip]{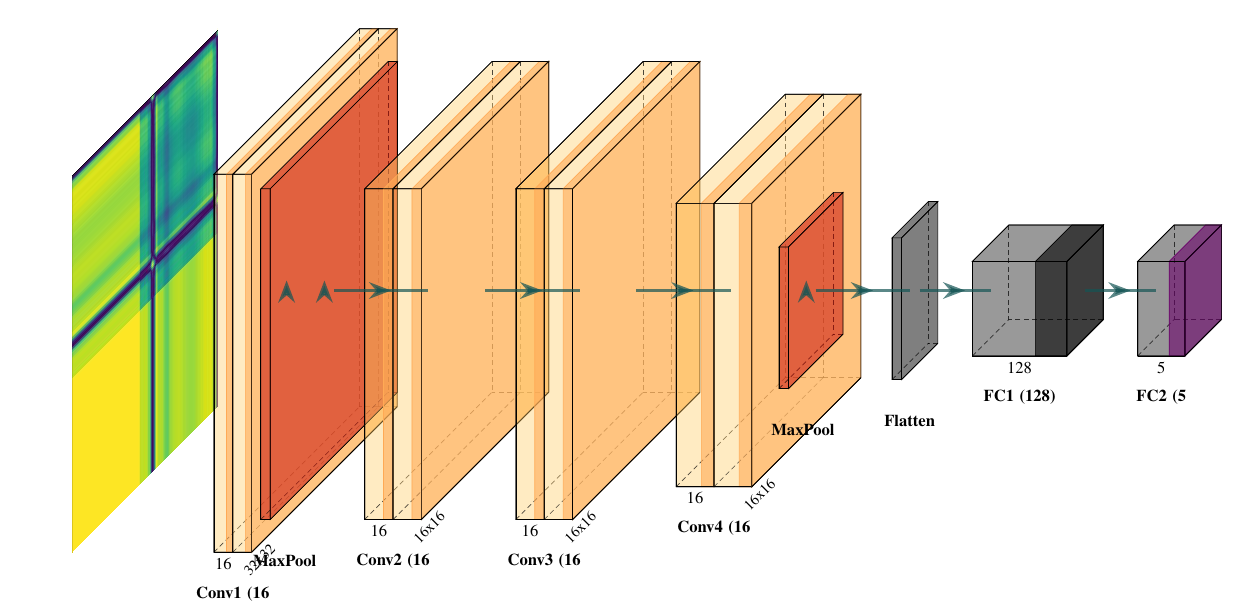}
    \caption{Architecture Diagram for the proposed CNN model}
    \label{fig:ecg_cnn_architecture}
\end{figure*}

The 2D Convolutional Neural Network (CNN) architecture used for ECG classification, depicted in Figure \ref{fig:ecg_cnn_architecture}, comprises several key layers designed for effective feature extraction and classification:

The CNN architecture was adapted from \cite{elmir2023ecg} and optimized through cross-validation to balance computational load and performance on low-power devices. Kernel sizes (7×7 and 5×5) were empirically selected to preserve morphological ECG features while maintaining low inference latency.

\begin{itemize} 
\item \textbf{Convolutional Layers}: The network includes four convolutional layers. The first layer has a 7x7 kernel with padding, followed by three additional layers with 5x5 kernels. Each layer applies LeakyReLU activations, and two max-pooling layers are included after the first and fourth convolutional layers to progressively reduce spatial dimensions. These layers extract spatial features critical for ECG pattern recognition. 
\item \textbf{Fully Connected Layers}: After flattening, the output from the convolutional layers is passed through a fully connected layer with 128 neurons, which further condenses the spatial features. 
\item \textbf{Output Layer}: A softmax layer with five output neurons classifies the ECG data into five distinct categories, producing probabilistic outputs across classes. 
\end{itemize}

This architecture is optimized for the GAF-transformed images of ECG, leveraging spatial feature extraction through CNNs to enable accurate classification on both high-capacity and low-capacity devices.

This architecture is implemented on both the server and clients, with each client independently updating the model using its local data.

\section{Experimental Setup}
\subsection{Dataset and Preprocessing}

The MIT-BIH Arrhythmia dataset \cite{moody2001impact} is widely used for ECG signal classification and arrhythmia detection. In line with previous studies \cite{kiranyaz2015real, li2017classification}, we collected a total of 26,490 samples, categorized into five heartbeat types for classification: N (normal beat), L (left bundle branch block), R (right bundle branch block), A (atrial premature contraction), and V (ventricular premature contraction). Half of the samples were randomly selected for training, while the remaining samples were reserved for testing.

The dataset was partitioned across clients to simulate realistic IoT healthcare scenarios, where ECG data are naturally distributed across hospitals, personal monitoring devices, or wearable sensors. This heterogeneous distribution ensures that training conditions reflect real-world non-centralized environments, where each device retains ownership of its local data to preserve privacy. The dataset is partitioned to reflect real-world deployment scenarios: the Raspberry Pi, simulating a wearable device, receives 1\% of the dataset; the laptop, representing emergency teams, receives 49\%; and the server, functioning as a local workstation, handles the remaining 50\%. Each client performs local training for ten epochs per federated round, after which the server aggregates model weights across ten rounds. The hardware setup includes a high-performance workstation as the server, with clients consisting of a laptop and a Raspberry Pi 4, the latter limited to CPU-based training to accommodate its resource constraints. Performance metrics include classification accuracy, communication overhead, and system performance indicators such as CPU usage and memory on the Raspberry Pi.

The ECG dataset is preprocessed by transforming 1D ECG signals into 2D images using Gramian Angular Field (GAF) representation \cite{elmir2023ecg}. The GAF transformation captures temporal correlations in a polar coordinate system, enabling 2D Convolutional Neural Networks (CNNs) to leverage spatial features of ECG patterns. Each signal is resized to 32x32 GAF images to maintain consistency.

\subsection{System Specifications}

The experiments were conducted on a Raspberry Pi 4, an Intel® Core™ i5-based server, and an Intel® Core™ i7 laptop, providing a diverse range of computational capacities for evaluating FL performance under realistic edge–cloud conditions. More details are presented in Table  \ref{table:hardware_comparison}.

\begin{table*}[!htbp]
\centering
\caption{Hardware Specifications Comparison}
\label{table:hardware_comparison}
\begin{tabular}{|l|p{4.5cm}|p{4.5cm}|p{4.5cm}|}
\hline
 \textbf{Specification} & \textbf{Raspberry Pi 4 model B} & \textbf{Intel® Core™ i5 Server} & \textbf{Intel® Core™ i7 Laptop} \\ \hline \textbf{Processor} & Broadcom BCM2711, Quad-core Cortex-A72 & Intel® Core™ i5-3570, Quad-core & Intel® Core™ i7-1165G7, 4-core \\ \hline
\textbf{Base Frequency} & 1.5 GHz & 3.4 GHz (Turbo Boost up to 3.8 GHz) & 2.8 GHz (Turbo Boost up to 4.4 GHz) \\ \hline
\textbf{Cache} & - & 6 MB (Intel® Smart Cache) & 12 MB (Intel® Smart Cache) \\ \hline
\textbf{Memory} & 8 GB & 12 GB & 12 GB \\ \hline
\textbf{Networking} & Gigabit Ethernet, 802.11ac Wi-Fi & Ethernet & 802.11ac Wi-Fi \\ \hline
\textbf{Power Efficiency} & High (5-10W) & Moderate (77W TDP) & Moderate (15-28W TDP) \\ \hline
\end{tabular}
\end{table*}

The devices vary significantly in processing power, memory, and power efficiency, which impacts their suitability for federated learning:
\begin{itemize}
    \item \textbf{Processing Power}: The Intel® Core™ i7-1165G7 in the laptop has a higher base and turbo frequency than the i5-3570 and Raspberry Pi, making it more efficient for compute-heavy operations.
    \item \textbf{Memory Capacity}: The laptop has 12 GB, the Raspberry Pi has 8 GB, and the server with 12 GB. Higher memory capacity benefits models requiring substantial in-memory computation.
    \item \textbf{Power Efficiency}: The Raspberry Pi, while limited in computational power, is highly energy-efficient and well-suited for distributed, low-power environments.
\end{itemize}

These hardware differences are essential for assessing the performance, resource usage, and practicality of federated learning on heterogeneous devices.

Each training experiment was repeated five times, and average accuracy with standard deviation was reported. Model training used a learning rate of 0.001, batch size of 32, and Adam optimizer. Class imbalance was managed by stratified sampling across clients to ensure representation consistency.

\section{Results \& Discussion}

\begin{table*}[!htbp]
\centering
\caption{Comparison of Single-Client (Experiment 1) and Multi-Client (Experiment 2) Results}
\begin{tabular}{|l|p{3.5cm}|p{3.5cm}|}
\hline
\textbf{Metric}                & \textbf{Experiment 1 \newline (Server only)} & \textbf{Experiment 2 \newline (Multi-Client)} \\ \hline
Training Time                  & 8518.64 sec                           & 5360.07 sec                          \\ \hline
Total Send Size                & 6116327 bytes                         & 18348981 bytes                        \\ \hline
Total Receive Size             & 6116285 bytes                         & 18348605 bytes                        \\ \hline
Train Accuracy                 & 87.81\%                               & 95.79\%                               \\ \hline
Test Accuracy                  & 87.30\%                               & 95.18\%                               \\ \hline
Accuracy by Class \newline (N, L, R, A, V) & 80\%, 96\%, 98\%, 48\%, 90\%       & 96\%, 95\%, 99\%, 82\%, 95\%          \\ \hline
\end{tabular}
\label{table:comparison}
\end{table*}

The results from two experimental setups—single-client (server only) and multi-client (server, Raspberry Pi 4, and laptop)—are summarized in Table~\ref{table:comparison}. These metrics cover key performance indicators including training time, communication overhead, and classification accuracy.

The results demonstrate that federated learning with GAF-transformed ECG signals can achieve high accuracy, even in a heterogeneous setup involving low-power devices. In Experiment 1, the single-client setup achieved a test accuracy of \textbf{87.30\%}, while the multi-client setup improved test accuracy to \textbf{95.18\%}.

Interestingly, the multi-client federated setup achieved a \textbf{shorter overall training time} (5360.07 sec) than the single-client scenario (8518.64 sec). This suggests that distributing the training workload across multiple devices, including resource-constrained platforms like the Raspberry Pi, enhances processing efficiency through parallel execution and distributed workload management—an effect similarly observed by Gao et al. \cite{gao2020end} in FL deployments on IoT systems.

As expected, the multi-client scenario incurred a higher total send and receive size due to increased model synchronization rounds between clients and the server, confirming trends reported by Jimenez Gutierrez et al. \cite{jimenez2023application} and Eleftheriadis and Karakonstantis \cite{eleftheriadis2024energy}. However, this increased communication overhead was offset by superior accuracy and improved training time, reinforcing the practical trade-off viability for FL in edge healthcare settings.

Building upon our previous work \cite{elmir2023ecg}, which validated GAF’s discriminative power in centralized ECG classification achieved up to \textbf{97.47\% accuracy and 98.65\% F1-score} in centralized anomaly detection tasks, our current multi-client FL implementation confirms that \textbf{GAF transformations remain effective for privacy-preserving, distributed ECG classification}. By adopting a uniform \textbf{32×32 image size}, we ensure compatibility with lightweight CNN models, mitigating computational load on devices like Raspberry Pi without sacrificing classification performance — an essential step for advancing GAF’s applicability in distributed, real-time ECG analysis.

The observed class-wise performance also aligns with previous literature on class imbalance and non-IID data challenges in FL ECG classification \cite{ccelik2023comparison, diao2020heterofl}. While overall per-class accuracy was strong in the multi-client setup, class “A” (atrial premature contraction) classification remained lower (82\%), suggesting a need for dataset rebalancing, personalized FL updates, or specialized model adjustments for minority classes.

In summary, this study confirms the feasibility and benefit of integrating federated learning with GAF-transformed ECG signals on heterogeneous, privacy-preserving, edge-capable platforms. The performance improvements, efficiency gains, and class-specific insights observed here complement and extend findings in recent FL and GAF literature \cite{elmir2023ecg, alsalemi2023lightweight, gao2025efficient}, further supporting this approach’s potential for scalable, real-world distributed healthcare systems.

These findings are particularly relevant in IoT healthcare scenarios, where communication bandwidth and energy efficiency are critical constraints. By demonstrating strong performance even on a Raspberry Pi, the framework shows promise for deployment on wearable IoT devices and smart health monitoring systems, extending scalability to broader edge-to-cloud IoT environments.

An additional ablation test excluding the Raspberry Pi client yielded 94.6\% accuracy, confirming that the inclusion of the IoT device contributed to distributed learning efficiency without degrading accuracy. The heterogeneous hardware contributed approximately 6\% of total training time variation.

Although our framework achieved competitive accuracy, future work will focus on enhancing the efficiency and scalability of federated ECG classification by optimizing FL aggregation strategies and integrating metaheuristic-based client optimization (e.g., Cuckoo Search) to improve convergence and communication performance. In addition, adaptive compression techniques \cite{gao2025efficient} will be explored to further reduce communication costs, while personalized FL strategies \cite{diao2020heterofl} will help mitigate non-IID data challenges. Finally, the use of hardware accelerators and quantized model optimization will be investigated to enable real-time deployment on ultra-low-power edge devices.

\section{Conclusion}
This work validates the practicality of combining Federated Learning (FL) with Gramian Angular Field (GAF) transformations for privacy-preserving electrocardiogram (ECG) classification across heterogeneous devices. By transforming 1D ECG signals into 2D GAF representations, the proposed FL-GAF framework leverages Convolutional Neural Networks (CNNs) for spatially enriched analysis while safeguarding data privacy and optimizing edge resource utilization. Experimental results show that the multi-client FL configuration attains a classification accuracy of 95.18\%, surpassing the single-client baseline in both accuracy and training efficiency. Notably, the framework sustains high performance on constrained platforms such as the Raspberry Pi, confirming its suitability for distributed IoT-based healthcare systems. Overall, the study highlights the promise of FL with GAF for scalable, secure, and efficient ECG classification in edge–cloud healthcare environments. Future efforts will aim to minimize communication overhead through adaptive compression, enhance robustness to non-IID data via personalized FL, and exploit hardware acceleration for real-time operation. Extending evaluations to diverse ECG datasets will further strengthen generalization across patient populations, advancing the development of practical, privacy-preserving IoT healthcare analytics.

\section*{Acknowledgment}

This work was supported by the PRFU project titled \textit{“Un système de santé intelligent et sécurisé pour la surveillance, la prédiction et la détection des maladies”}  under grant number \textbf{C00L07ES060120230002}.

\bibliographystyle{IEEEtran}
\bibliography{references}

\end{document}